\documentclass[runningheads]{llncs}
\usepackage{graphicx}
\usepackage{adjustbox}

\usepackage{caption}

\usepackage{enumitem}
\usepackage{cite}
\usepackage{xcolor}
\usepackage{amsmath,amssymb,amsfonts,euscript, mathrsfs}
\usepackage{textcomp}

\usepackage{hyperref}

\usepackage{algorithm} 
\usepackage{algpseudocode} 
\usepackage{subcaption}
\usepackage[nolist]{acronym}
\usepackage{xcolor}

\begin{document}
\title{subMFL: Compatible subModel Generation for Federated Learning in Device Heterogeneous Environment}
\titlerunning{subMFL: Compatible subModel Generation in FL}
%
%
\author{Zeyneddin Oz\inst{1}\orcidID{0000-0002-4216-9854} \and
Ceylan Soygul Oz\inst{3}\orcidID{0000-0001-6361-821X} \and
Abdollah Malekjafarian\inst{2}\orcidID{0000-0003-1358-1943} \and
Nima Afraz\inst{1}\orcidID{0000-0002-8422-9878} \and
Fatemeh Golpayegani\inst{1}\orcidID{0000-0002-3712-6550}}
\authorrunning{Z. Oz et al.}

%
\institute{University College Dublin, School of Computer Science, Belfield, Dublin, Ireland \and
University College Dublin, School of Civil Engineering, Belfield, Dublin, Ireland \and
DOCOsoft, NexusUCD Belfield Office Park, Dublin, Ireland}
\maketitle              
\begin{abstract}
Federated Learning (FL) is commonly used in systems with distributed and heterogeneous devices with access to varying amounts of data and diverse computing and storage capacities. FL training process enables such devices to update the weights of a shared model locally using their local data and then a trusted central server combines all of those models to generate a global model. In this way, a global model is generated while the data remains local to devices to preserve privacy. However, training large models such as Deep Neural Networks (DNNs) on resource-constrained devices can take a prohibitively long time and consume a large amount of energy. In the current process, the low-capacity devices are excluded from the training process, although they might have access to unseen data. To overcome this challenge, we propose a model compression approach that enables heterogeneous devices with varying computing capacities to participate in the FL process. In our approach, the server shares a dense model with all devices to train it: Afterwards, the trained model is gradually compressed to obtain submodels with varying levels of sparsity to be used as suitable initial global models for resource-constrained devices that were not capable of train the first dense model. This results in an increased participation rate of resource-constrained devices while the transferred weights from the previous round of training are preserved. Our validation experiments show that despite reaching about 50 per cent global sparsity, generated submodels maintain their accuracy while can be shared to increase participation by around 50 per cent.
\vspace{-0.3cm}
\keywords{Resource-constrained heterogeneous edge devices \and Federated learning \and Model pruning \and Mobile edge devices.}
\end{abstract}
%
%
%

\vspace{-1.1cm}
\section{Introduction}
\vspace{-0.3cm}
\subsection{Background}
\vspace{-0.3cm}
The widespread use of smart devices like smartphones, tablets, and Internet of Things (IoT) devices of various sizes and purposes, is driving the progress of services in smart environments, including smart cities, intelligent transport systems and infrastructure \cite{b1, b2, b3}. Furthermore, the massive quantity of edge devices is expected to generate extensive data requiring processing and analysis through automated methods. Machine learning can fuel the emergence of novel applications in smart environments by using those data \cite{b4}. Smart cities and their associated services such as intelligent traffic management, waste management, surveillance, and infrastructure monitoring are examples of such environments.

The exponential growth of generated data by IoT and mobile devices, along with demands for low latency computation, privacy, and scalability, drives the shift to edge computing. This approach enhances model training by placing computation nearer the data source hence reducing the data transmission latency. However, edge devices (i.e., edge nodes) often have limited computation power, storage, and energy capacity, making it challenging to run computationally intensive applications, mainly when a large amount of data must be processed \cite{b5}.
\vspace{-0.6cm}
\begin{figure}[htp]
    \centering
    \includegraphics[trim={0 0.75cm 0 1cm}, clip, width=0.8\textwidth]{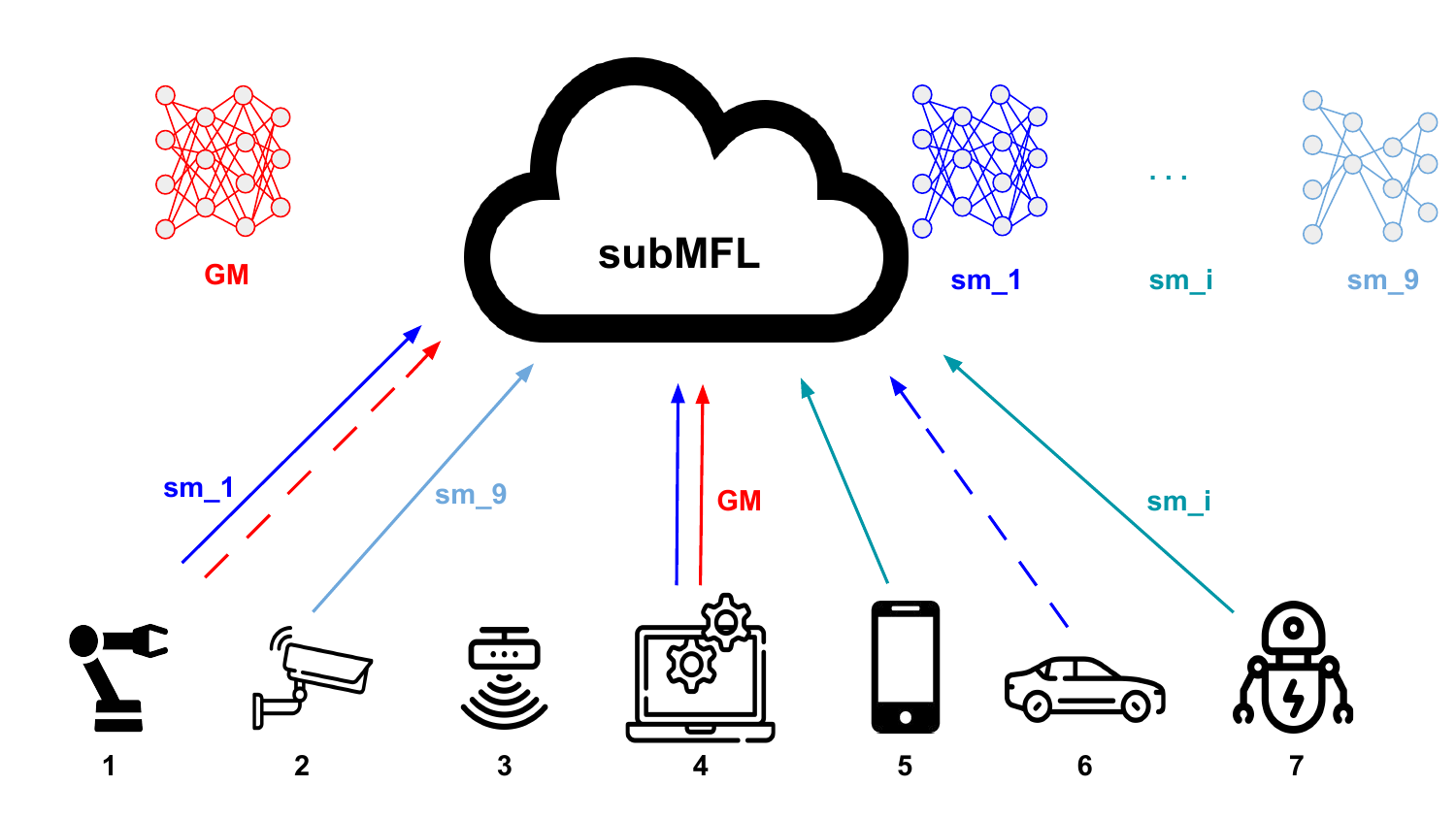}
    \caption{A representation of subMFL working with a dense global model ($GM$) and generated submodels ($SM = [{sm_1, sm_2, ... , sm_9}]$).}
    \label{AAA}
\end{figure}
\vspace{-0.8cm}

Distributed machine learning refers to multi-node machine learning algorithms and systems that are designed to improve performance, increase accuracy, and scale to larger input data sizes \cite{b6}. Powerful parallel and distributed computing systems have recently become widely accessible in multi-core processors and cloud computing platforms that are applicable to problems traditionally addressed by centralised and sequential approaches \cite{b7}. Standard distributed learning involves training Deep Neural Networks (DNNs) on cloud servers and deploying them to edge nodes. However, this will not perform well for applications needing low latency, privacy, and scalability. Centralized model training demands data sharing, however, this may discourage data owners from granting access to their data for the purpose of model training.

In such a setting, machine learning models must be trained either at the same nodes that generate them (also can be defined as an agent, client, worker or device) or at a set of intermediate nodes, each collecting a subset of the data. Federated Learning (FL) \cite{b8} enables distributed machine learning across a large number of devices without requiring them to share their data with a central server. Once the devices train their local model using the devices' local model parameters it is returned to the central servers to be aggregated with other sub-models and get distributed to all devices. A key challenge in deploying FL is the vast heterogeneity of devices \cite{b9}, ranging from low-end IoT e.g., humidity sensors to mobile devices, as shown in Fig. 1, each having access to various types and amounts of data and hardware.
\vspace{-0.7cm}
\begin{figure}
    \center
    \includegraphics[trim={0 3.8cm 0 4.05cm}, clip, width=0.9\textwidth]{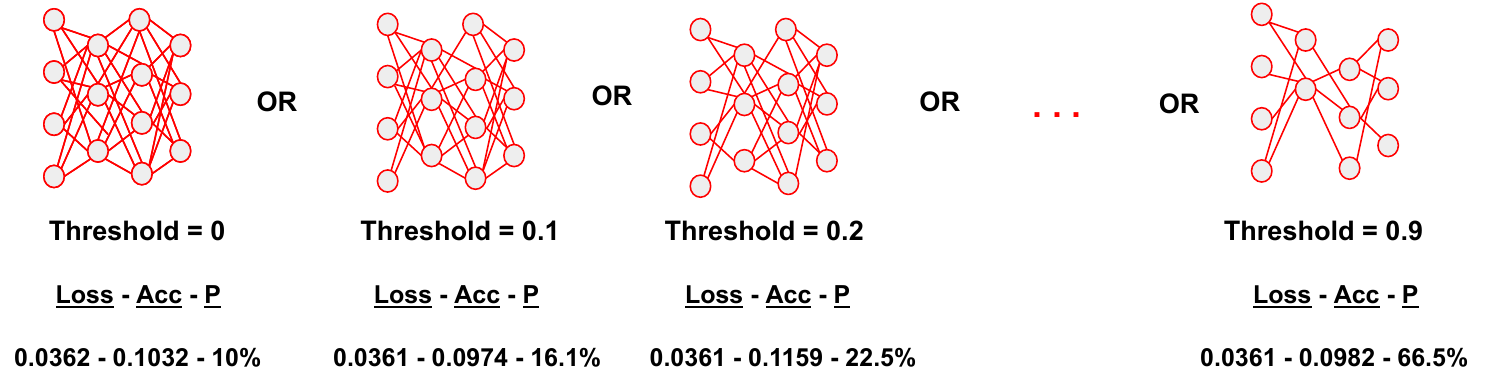}
    \caption{Standard Federated Learning system's initial global models' performance.}
    \label{AAA}
\end{figure}
\vspace{-0.85cm}

The widely accepted approach in FL requires all devices to use the same global model. However, this causes a problem when large-scale models such as DNNs with a large number of parameters must be used by resource-constrained devices. To use dense DNNs in FL systems, developers frequently choose to exclude such devices from training, which results in training bias and affects the model generality, due to excluding the data that was owned by such devices \cite{b10}. Another approach is to reduce the global model's size by its depth or width, to accommodate the resource-constrained devices. However, this results in lower accuracy due to model capacity constraints \cite{b11}. Fig. 2 shows a representation of the density range of possible initial global models by the width (dropped neurons or links) that own randomly generated weights in the Standard Federated Learning (S-FL). Picking one of the models arbitrarily as a global model to share with all devices to train leads to a tradeoff between participation rate and model learning constraints, due to the size of the selected model. While randomised model selection leads to this issue, DNNs pruning has the potential to generate sparse and suitable models. For instance, to utilise in FL, this compression technique is proposed to generate purposefully sparse models to address challenges such as communication overhead \cite{b12}, data heterogeneity \cite{b13}, and inclusion of heterogenous devices \cite{b14}. 

\vspace{-0.5cm}
\subsection{Related Works}
\vspace{-0.2cm}
Looking at the literature in more detail, FedSCR \cite{b12} reduces upstream communication by clustering parameter update patterns and using sparsity through structure-based pruning. Hermes \cite{b15} uses structured pruning to find small subnetworks on each device, and only updates from these subnetworks are communicated, improving communication and inference efficiency. AdaptCL \cite{b16} sends different stage pruned models to each device to synchronize the FL process in a heterogeneous environment and converge device update response times. 

To address data heterogeneity, some methods cluster devices based on parameters and aggregate each cluster's parameters separately \cite{b13}. In \cite{b17} a custom pruning was introduced that maximizes the coverage index via utilising a local pruning mask, considering both pruning-induced errors and the minimum coverage index, instead of solely preserving the largest parameters. 

Considering the pruning in terms of device heterogeneity, in PruneFL \cite{b18}, first an initial device is selected to prune the initial global model, and then further pruning involves both the server and devices during the FL process. Thus, a good starting point can be found for the FL process involving all devices. In \cite{b19}, before the FL process, it is suggested to run local dataset-aware pruning, in order to achieve device-related models. In \cite{b20}, it is proposed that during FL training, the server determines a pruning ratio and allocates wireless resources adaptively. Then, a threshold-based device selection strategy is used to further improve the learning performance. The approach of FedPrune \cite{b21} is to send randomly different sub-models to devices from the server side to find the optimum sub-model. FedMP \cite{b22} enables each device to avoid training the entire global model by determining specific pruning ratios. In FjORD \cite{b14}, system diversity is considered and the same model size is not shared to all devices, instead, the model size is tailored to the devices. 

\vspace{-0.4cm}
\subsection{Motivations and Contributions}
\vspace{-0.2cm}
There have been works investigating the implementation of DNNs pruning in FL. However, there is a need for determining a specific pruning ratio or the computation cost of the pruning process is left to the device side in existing studies that address device heterogeneity. This results in extra energy costs until finding the optimum trainable model architecture due to over pruning process on such resource-constrained devices. 

In this paper, we focus on developing a novel model that enables heterogeneous devices to participate in the FL process by proposing a Compatible subModel Generation in FL (subMFL). subMFL aims to produce suitable submodels considering their initial accuracy despite their smaller size, instead of randomly generated smaller models in S-FL. In this model, an initial dense Global Model ($GM$) with low learning constraints is shared with all devices and it is trained by devices with enough resources. When training is completed, the model is pruned dataless to generate compatible submodels to be used by resource-constrained devices, without a need for prior knowledge of their computation and communication capabilities or determining a specific pruning ratio. For evaluation purposes, different threshold values are used to generate submodels with different sparsity levels. The accuracy of models and the level of participation of devices for each set are then reported. Our main contributions in this paper are as follows:
\vspace{-0.2cm}
\begin{itemize} 
\item\textbf{Server-side model pruning:} The over-pruning process on the device side leads to extra energy loss. In our work, the pruning stage is completely carried to the server without the need for any data sample.  

\item\textbf{Compatible subModels generation:} By assuring that the trained dense model's weights are transferred to generated submodels, resource-constrained devices benefit from the data used to train $GM$ by beginning to train a model with satisfactory accuracy.

\item\textbf{Increased heterogeneous devices participation rate:}  subMFL tailors the FL paradigm, for environments that include heterogeneous devices with various levels of computational resources by assigning suitable pre-trained and compressed initial global models that fit their resources (Fig. 1).
\end{itemize}

\vspace{-0.8cm}
\section{Compatible subModel Generation in Federated Learning}
\vspace{-0.4cm}
In this paper, we propose a Compatible subModel Generation model to enhance Federated Learning (subMFL) in environments with heterogeneous resource constraint devices with varying computational capacities. subMFL uses pruning to generate a set of compatible sparsed submodels using a trained dense Global Model ($GM$). Those will be initial global model architectures with suitable size that allows resource-constrained devices to join in the upcoming training cycles.
\vspace{-0.95cm}
\begin{algorithm}
	\caption{subMFL: Compatible subModel Generation in Federated Learning} 
	\begin{algorithmic}[1]
        
        \State Server generates $GM$ architecture
        \State $\mathit{W}_{GM}$ = \Call{TrainModelOnDevices}{$\mathit{W}_{GM}, D, T$} 
        \State $SM$ = \Call{GenerateSubModels}{$\mathit{W}_{GM}$}
        \For{each $\mathit{W}_{sm_i}$ in $SM$}
            \State $D$ = \Call{DropDevices}{$\mathit{W}_{sm_i}, D$}
            \State ${W}_{sm_i}$ = \Call{TrainModelOnDevices}{$\mathit{W}_{sm_i}, D, T$}
        \EndFor
	\end{algorithmic} 
\end{algorithm}
\vspace{-0.8cm}

\textbf{subMFL Flow:} Algorithm 1 shows the overall flow of subMFL. The stages are: A dense global model is generated in the server and distributed to all devices to be trained ($T$ represents the global training round and we set it as 100 in our simulation. $W = [w_0, w_1, \ldots, w_n], \quad 0 \leq w_i \leq 1 \quad \text{for } 0 \leq i \leq n$ represents weights and ${W}_{GM}$ are weights of $GM$). $GM$ is trained with capable devices and then a set of sparsed submodels ($SM = [{sm_1, sm_2, ... , sm_9}]$) is generated by pruning this $GM$ using different threshold values. Afterwards, $SM$ will be sent to all devices starting from the densest to the sparsest submodel, and each device chooses to train the densest compatible submodel based on its computational resources. At each step, the devices themself make a decision to join the next round based on their local model accuracy. Devices that reach their preferred accuracy exit the training process, and the next submodel is not shared with them. We represent devices with $D = [{d_1, d_2, ... , d_{1000}}]$ and in line 5, we update this device set based on their preference. Thus, such devices do not consume further energy when the target accuracy is reached. Each component of subMFL is as follows:

\textbf{Training:} Algorithm 2 shows the training procedure we used, which is the process in standard federated learning (S-FL). Weights of the current global model are shared with $D$ to train with $T$ global round. At each round, local models ($\mathit{W}_{LMs}^t$) are collected and aggregated, to update the global model with new weights. Then, this updated ${W}_{GM}$ are shared with $D$ to be updated again with their local datasets. In device heterogeneous environments, each device needs a different amount of time to complete its local training which causes synchronisation issues. On the other hand, in subMFL, devices that are slow to train the current model already cannot send local updates, however, the model is trained with higher capacity devices. Including $GM$, there is $SM$ that will be trained and at each step of distributing a sparser submodel, devices that have near resource capacity train the distributed model, which leads the server to receive local models synchronously. In this way, we do not need pre-information about the devices' computation capacity or determine a specific pruning ratio.
\vspace{-0.7cm}
\begin{algorithm}
    \caption{Training Process:} 
	\begin{algorithmic}[1]
        \Function{TrainModelOnDevices}{$\mathit{W}_{M}, D, T$}
    		\For {each round $t=1,2,\ldots,T$}
                \State Server shares $\mathit{W}_{M}^{t}$ with devices $\mathit{D}$
                \State Server receives $\mathit{W}_{LMs}^t$ from devices $\mathit{D}$
                \State $\mathit{W}_{M}^{t}$ = ModelAggregate($\mathit{W}_{LMs}^t$) with FedAvg
            \EndFor  
            \State \Return $\mathit{W}_{M}^{t}$
        \EndFunction
    \end{algorithmic} 
\end{algorithm}
\vspace{-0.8cm}

For the aggregation process, we use the FedAvg algorithm \cite{b23}, which is an advanced aggregation strategy that has the benefits of convergence guarantees. This algorithm will be updated according to the current $SM$ architecture. 
\vspace{-0.7cm}
\begin{algorithm}
    \caption{Generating SubModels:} 
	\begin{algorithmic}[1]
        \Function{GenerateSubModels}{$\mathit{W}_{GM}$}
            \State $Threshold = 0$
            \State $SM = []$
            \While {$Threshold < 1$}
                \State $Threshold = Threshold + 0.1$
                \For {each $\mathit{w_i}$ in $\mathit{W}_{GM}$}
                    \If {$\mathit{w_i} < Threshold$}
                        \State $\mathit{w_i} = 0$
                    \EndIf
                \EndFor
            \State Add $\mathit{W}_{GM}$ to $SM$
            \EndWhile
            \State \Return $SM$
        \EndFunction
    \end{algorithmic} 
\end{algorithm}
\vspace{-0.6cm}

\textbf{Generating SubModels:}
DNNs pruning is used to generate the $SM$ from the $GM$ to be distributed to the resource-constrained devices. This will increase the participation of the heterogeneous devices that could not take part in the training process due to having a more limited resource capacity. In FL, due to security and privacy concerns, the server is unable to see any data sample which makes it unsuitable to prune DNNs on the server side with the majority of pruning methods. For this reason, we utilised a dataless pruning method on the server side, which is critical for real-world applications. In this way, all pruning processes are carried out at the server to decrease energy usage in resource-constrained devices. Also, we used an unstructured pruning strategy based on the L1-norm, due to its independence from network configuration \cite{b24}. 

Algorithm 3 shows submodels generation, where a $Threshold$ variable that ranges from 0 to 0.9, increasing 0.1 each time is defined for pruning the GM. In this process, the weights of the GM are below the selected threshold will be set to 0. The remaining weights will be transferred from the current global model to the newly pruned submodel. Since the threshold is incremented by 0.1, $GM$ produces 9 different submodels ($sm_i \in SM$) with various sparsification ratios. As shown in Fig. 3, $GM$ (see the red model) is the dense model and $SM$ (see the blue models) is generated using the pruned version of the trained $GM$.

\vspace{-0.6cm}
\begin{figure}
    \center
    \includegraphics[trim={0 2.8cm 0 0.44cm}, clip, width=\textwidth]{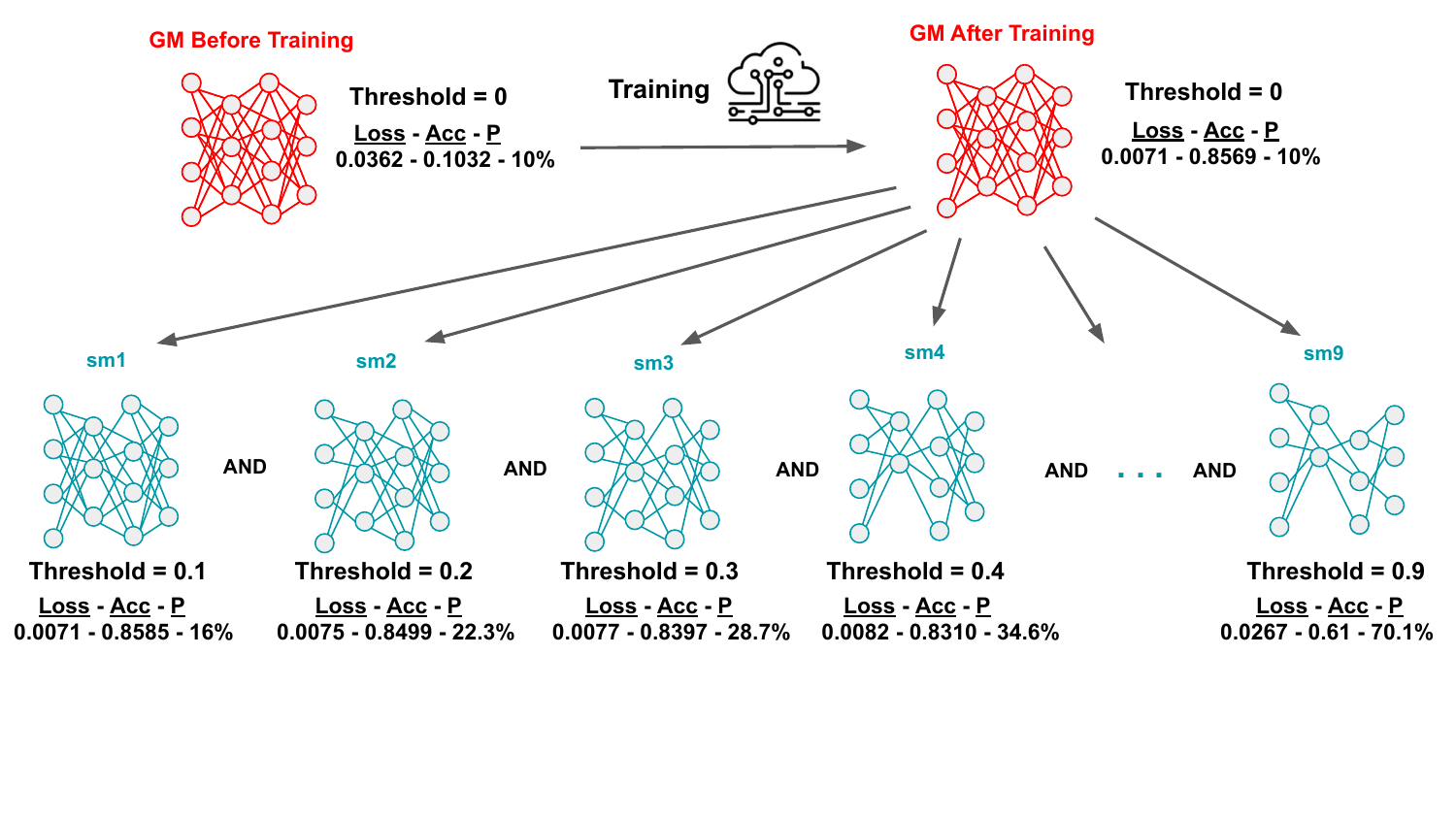}
    \caption{Loss, Accuracy and Participation performance of $SM$ using MNIST.}
    \label{AAA}
\end{figure}
\vspace{-0.75cm}

\textbf{Dropping Devices:} When devices reach their target accuracy they don't train the next $SM$ (Fig-1 represents the scenario, device-5 trains one of $sm_i$, but doesn't attend to train $sm_9$). For this reason, the server shares the next densest model only with devices that join the training. Algorithm 4 shows how the server updates $D$, which includes the devices that join the training. In line 3, ${d}_{jTargetMinAcc}$ shows minimum target accuracy for the device $d$. Following this approach, we reduce energy usage by omitting devices that reached the target.
\vspace{-0.7cm}
\begin{algorithm}
    \caption{Dropping Devices:} 
	\begin{algorithmic}[1]
        \Function{DropDevices}{$\mathit{W}_{sm_i}, D$}
            \For{each device $d_j$ in $D$}
                \If{${d}_{jTargetMinAcc} \leq \mathit{sm}_{iAcc}$}
                    \State Remove $d_j$ from $D$ 
                \EndIf
            \EndFor
            \State \Return $D$
        \EndFunction
    \end{algorithmic} 
\end{algorithm}
\vspace{-0.8cm}

As a result, instead of picking a random global model architecture generated with random weights as shown in Fig. 2, trained $GM$ can produce $SM$, and then those $SM$ can be shared to train with resource-constrained devices as sparser global models. In our approach, even though the next global models become smaller and have learning constraints due to compression, unlike S-FL, it keeps transferred weights from devices trained $GM$ and benefits from their unseen data. Thus, $GM$ is tuned to the available resource of devices, and devices can pick a $sm_i$. This way, resource-constrained devices aren't excluded from training.

\vspace{-0.6cm}
\section{Experiment}
\vspace{-0.3cm}
We used 1000 devices and shared data randomly with an equal sample size. 10 per cent of devices can train the dense global model ($GM$) and the remaining devices that have lower computational capacity train one of $sm_i$.

\textbf{Datasets:} Following the literature in this area, we used LeNet-5 \cite{b25} architecture with MNIST \cite{b26} and FMNIST \cite{b27} datasets which are used for image recognition tasks. While MNIST is a dataset of handwritten digits, FMNIST is a dataset of images depicting various clothing items. 

\textbf{Settings:} We have performed the simulation using Pytorch \cite{b28} and Flower \cite{b29} framework. The global round is set to 100 and the local epoch is 3. The validation data percentage is 10 and the batch size is 64. We used Adam \cite{b30} optimiser with a 0.001 learning rate. The remaining parameters are as follow: betas=(0.9, 0.999), eps=1e-08, weight decay=0, amsgrad=False, foreach=None, maximize=False, capturable=False, ”min fit clients” and ”min eval clients”=3.

\textbf{Availability:} In real-life scenarios, it is unlikely to receive parameters from all devices in every round due to factors such as mobility, low energy, and connection issues. Therefore, we assumed that only 30 per cent of devices are available. If this number is decreased, the convergence time of $GM$ increases.

\textbf{Baseline:} We used standard Federated Learning (S-FL) as our baseline.

\textbf{Evaluation metrics:} Our evaluation metrics include accuracy (Acc), loss (Loss), participation number (P), and global sparsity (GS) of S-FL and subMFL. Server-side threshold-based model pruning is used to generate $SM$. For instance, when the threshold is set as 0.1, parameters of $GM$ under 0.1 are reduced to 0, to generate the first submodel. To generate sparser submodels, the threshold increases until 0.9. By increasing the threshold, submodels become sparser, reducing computational cost and increasing the number of participating devices. We analyse metrics based on different threshold values and compare the results of subMFL generated $SM$ with sparse models in S-FL using the same thresholds. The code of this work is publicly available at: \href{https://github.com/zeyneddinoz/subMFL}{https://github.com/zeyneddinoz/subMFL} 

\vspace{-0.5cm}
\subsection{Results}
\vspace{-0.7cm}
\begin{table}[]
\centering
\begin{tabular}{c|c|c|c|c|c|c|c|c|c|}
 & \textbf{T} & \vtop{\hbox{\strut \textbf{S-FL}}\hbox{\strut \textbf{Acc}}} & \vtop{\hbox{\strut \textbf{subMFL}}\hbox{\strut \textbf{Acc}}} & \vtop{\hbox{\strut \textbf{S-FL}}\hbox{\strut \textbf{Loss}}} & \vtop{\hbox{\strut \textbf{subMFL}}\hbox{\strut \textbf{Loss}}} & \vtop{\hbox{\strut \textbf{S-FL}}\hbox{\strut \textbf{P}}} & \vtop{\hbox{\strut \textbf{subMFL}}\hbox{\strut \textbf{P}}} & \vtop{\hbox{\strut \textbf{S-FL}}\hbox{\strut \textbf{GS}}} & \vtop{\hbox{\strut \textbf{subMFL}}\hbox{\strut \textbf{GS}}} \\ \hline
\textit{\textbf{GM}}  & 0.0 & 0.1032 & 0.1032 & 0.0362 & 0.0362 & 100 & 100 & 0.0 & 0.0 \\
\textit{\textbf{sm1}} & 0.1 & 0.0974 & 0.8585 & 0.0361 & 0.0071 & 161 & 160 & 6.19 & 6.05 \\
\textit{\textbf{sm2}} & 0.2 & 0.1159 & 0.8499 & 0.0361 & 0.0075 & 225 & 223 & 12.57 & 12.38 \\
\textit{\textbf{sm3}} & 0.3 & 0.0935 & 0.8397 & 0.0361 & 0.0077 & 284 & 287 & 18.48 & 18.70 \\
\textit{\textbf{sm4}} & 0.4 & 0.1133 & 0.8310 & 0.0361 & 0.0082 & 353 & 346 & 25.39 & 24.66 \\
\textit{\textbf{sm5}} & 0.5 & 0.1135 & 0.8301 & 0.0361 & 0.0084 & 423 & 410 & 32.34 & 31.09 \\
\textit{\textbf{sm6}} & 0.6 & 0.0986 & 0.8422 & 0.0361 & 0.0081 & 478 & 475 & 37.83 & 37.56 \\
\textit{\textbf{sm7}} & 0.7 & 0.0823 & 0.8472 & 0.0361 & 0.0084 & 551 & 536 & 45.18 & 43.61 \\
\textit{\textbf{sm8}} & 0.8 & 0.0892 & 0.8380 & 0.0362 & 0.0110 & 614 & 607 & 51.40 & 50.78 \\
\textit{\textbf{sm9}} & 0.9 & 0.0982 & 0.6169 & 0.0361 & 0.0267 & 665 & 701 & 56.56 & 60.13 \\ \hline
\end{tabular}
\caption{Metrics values based on thresholds for MNIST dataset.}
\label{tab:my-table}
\end{table}
\vspace{-1.2cm}
Table 1 reports the results we obtained from different thresholds (T, e.g. 0.1 to 0.9) to generate 9 different submodels ($sm_i \in SM$, see Fig. 3) which includes accuracy, loss, number of participating devices (P) and global sparsity (GS) on generated models. To compare with S-FL, we picked models with different sparsification levels based on the same threshold values as shown in Fig. 2. In our experiments, the pruning method increases the sparsification of trained $GM$ significantly, while maintaining good accuracy in subMFL. Parallel to the increased model sparsity, the number of participating devices increases.

\textbf{Accuracy vs Global Sparsity:} As shown in Fig. 4-a and Fig. 4-c, the results show that when the threshold value is incremented global sparsity increases. However, independent from the sparsification of models, S-FL accuracy remains around 10 per cent, due to picked models always starting with randomly generated weights. On the other hand, in the beginning, the dense global model ($GM$) accuracy is the same as in S-FL, however, after training $GM$, generated submodels ($SM$) accuracy values are high. Thus, although the global model was sparsed, the transferred weights from previous training allowed the model to maintain a good level of accuracy. For instance, even though when global sparsity increases by around 50 per cent, the accuracy decreases by only about 2 per cent for the MNIST dataset. For the same condition, the accuracy percentage decreases approximately by 10, for the FMNIST dataset.
\vspace{-0.8cm}
\begin{figure*}
 \begin{subfigure}{0.49\textwidth}
     \includegraphics[trim={0 6.5cm 0 6.5cm}, clip, width=0.9\textwidth]{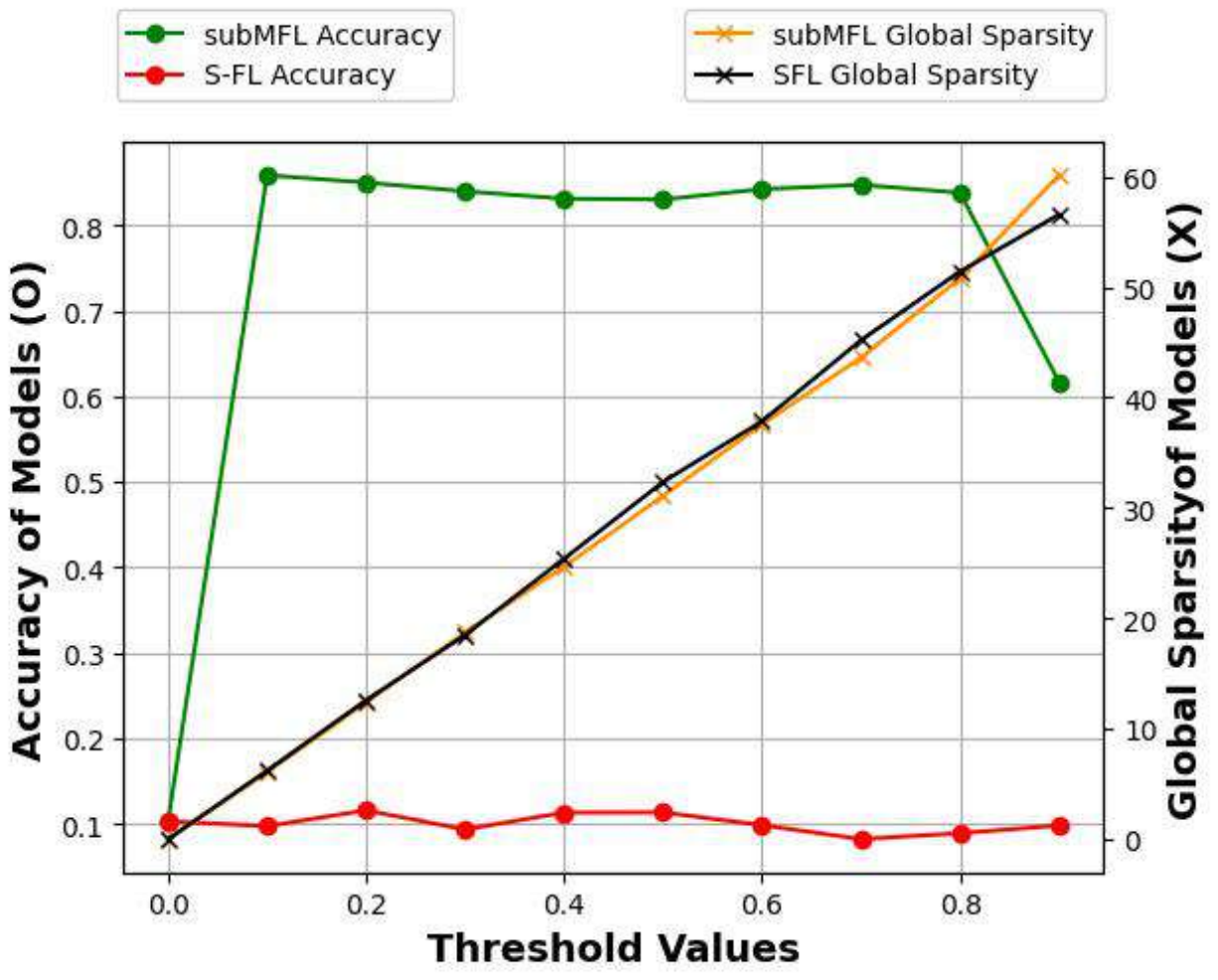}
     \caption{Accuracy vs Global Sparsity Percentage with MNIST benchmark.}
 \end{subfigure}
 \hfill
 \begin{subfigure}{0.49\textwidth}
     \includegraphics[trim={0 6.5cm 0 6.5cm}, clip, width=0.9\textwidth]{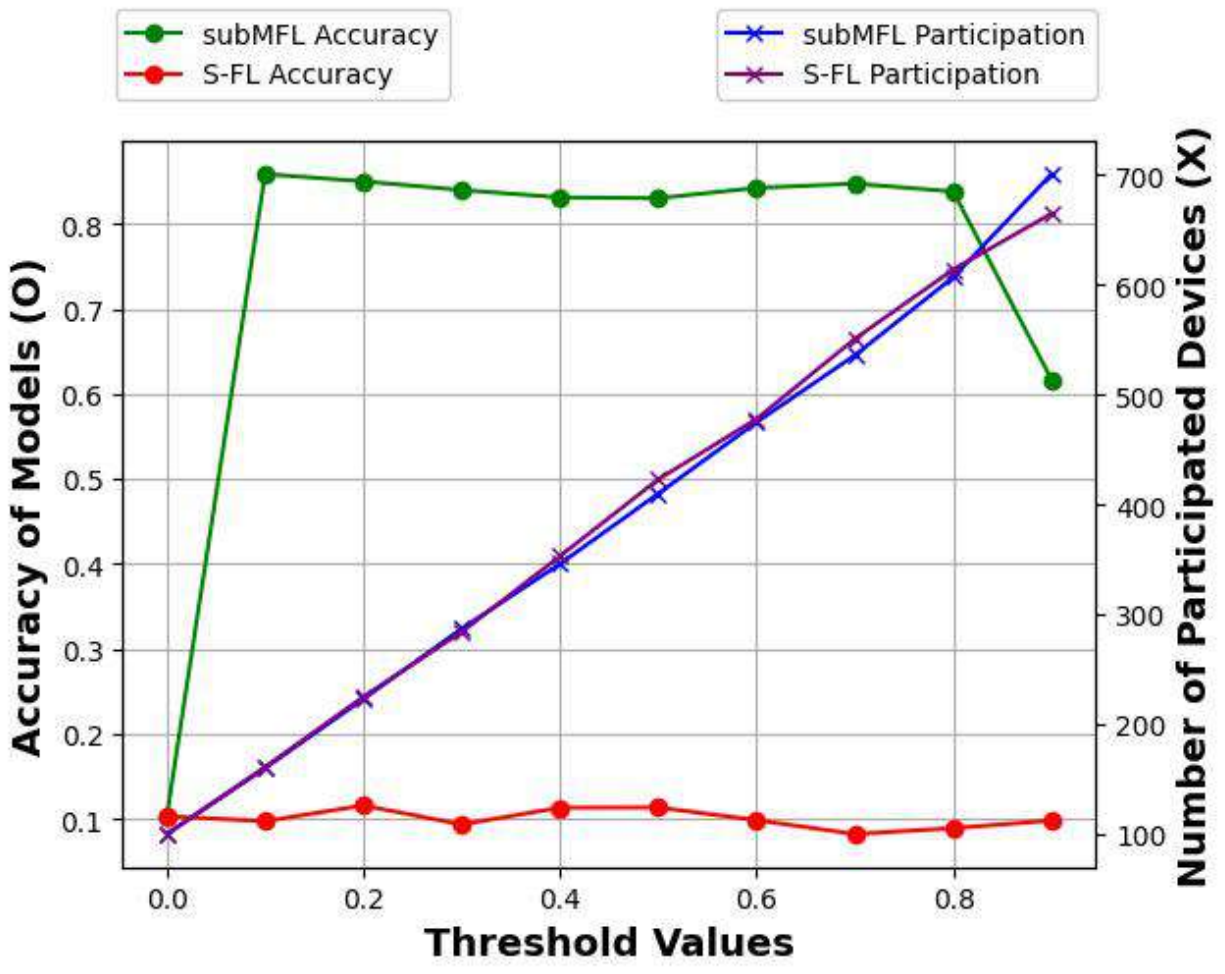}
     \caption{Accuracy vs Number of Participating Devices with MNIST benchmark.}
 \end{subfigure}
 
 \medskip
 \begin{subfigure}{0.49\textwidth}
     \includegraphics[trim={0 6.5cm 0 6.5cm}, clip, width=0.9\textwidth]{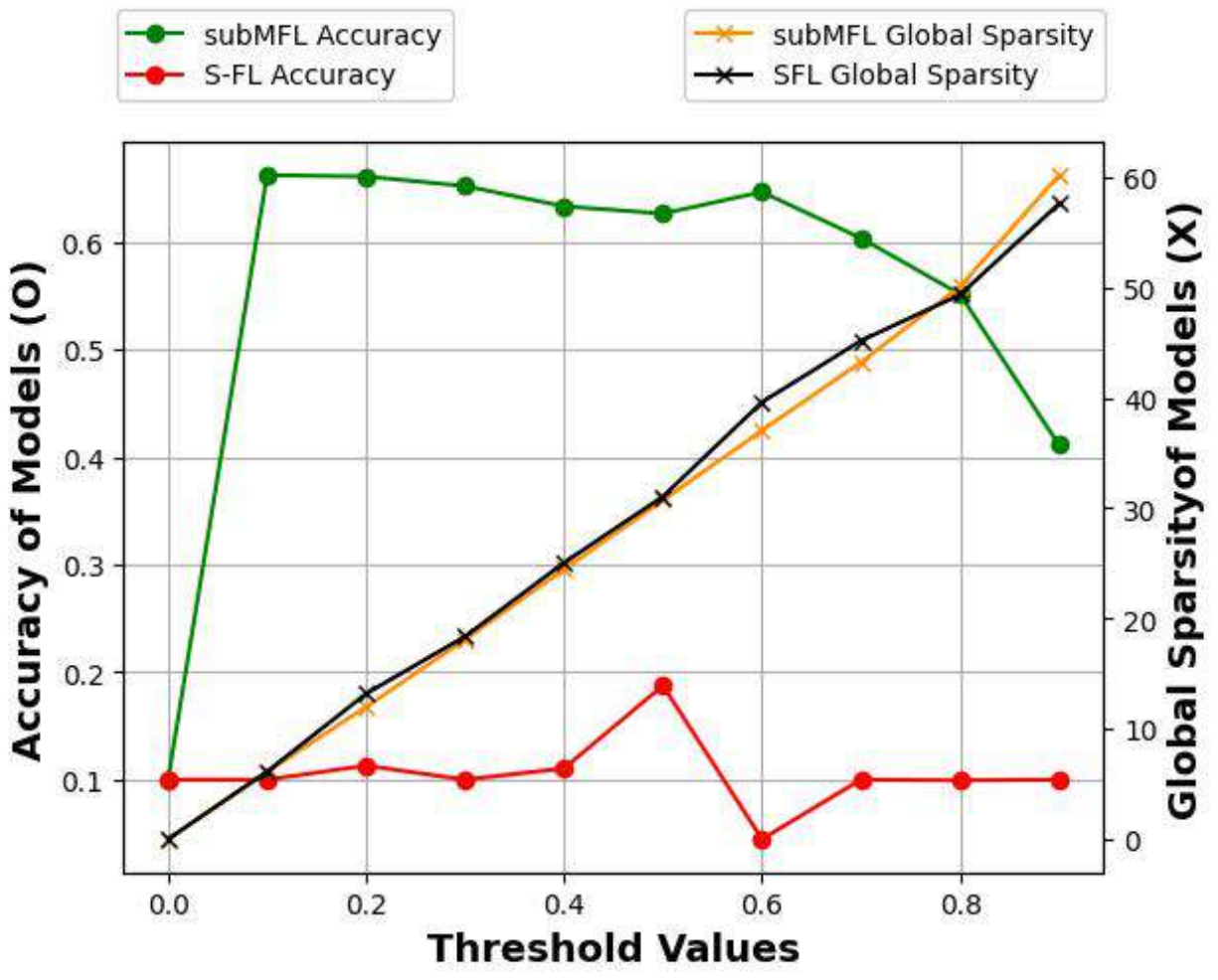}
     \caption{Accuracy vs Global Sparsity Percentage with FMNIST benchmark.}
 \end{subfigure}
 \hfill
 \begin{subfigure}{0.49\textwidth}
     \includegraphics[trim={0 6.5cm 0 6.5cm}, clip, width=0.9\textwidth]{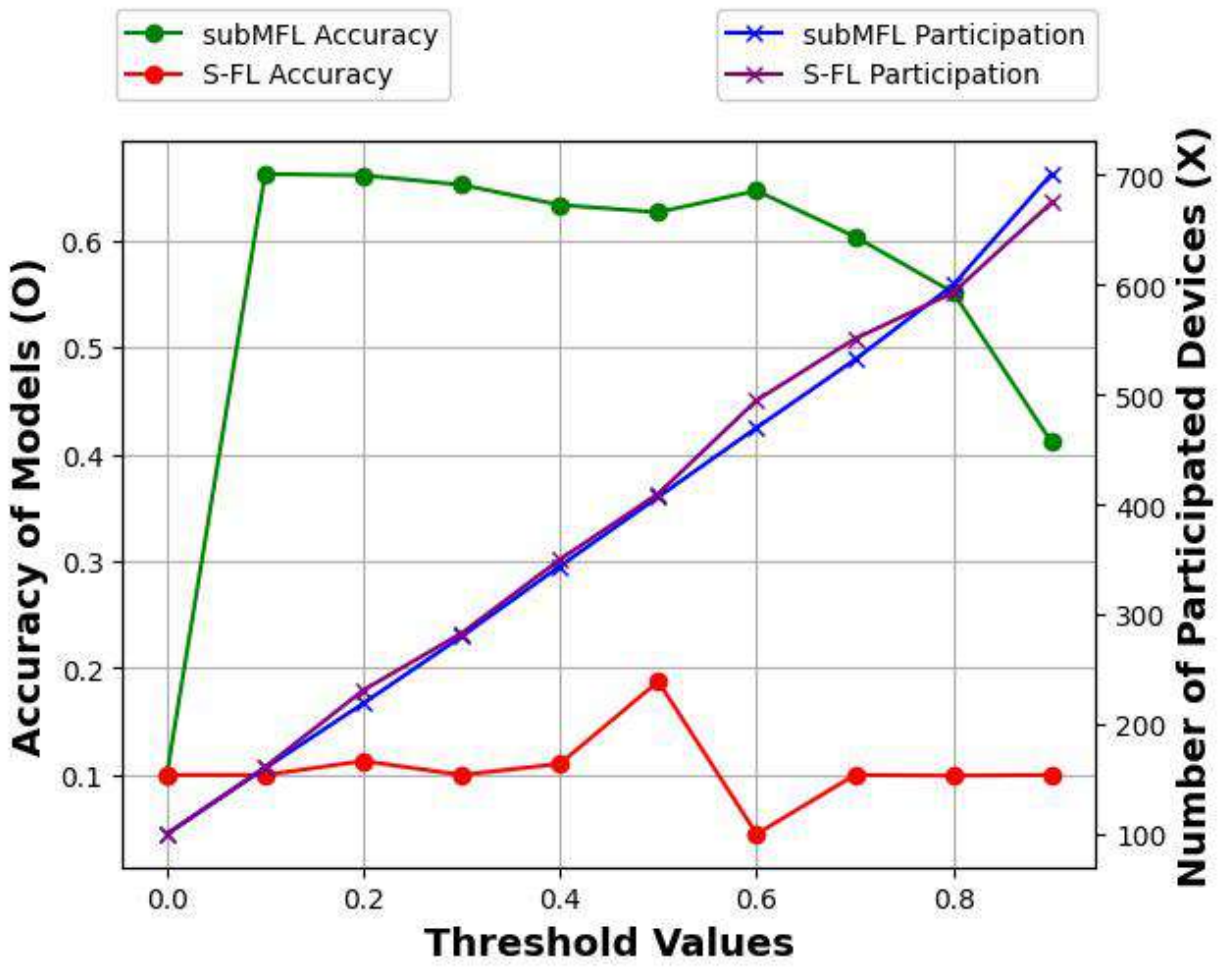}
     \caption{Accuracy vs Number of Participating Devices with FMNIST benchmark.}
 \end{subfigure}

 \caption{Accuracy vs Global Sparsity and Accuracy vs Participation.}
 \label{Label}
\end{figure*}
\vspace{-0.84cm}

\textbf{Accuracy vs Participation:} As a result of compression, resource-constrained devices can train compressed submodels, and the participation number increases (see Fig. 4-b and Fig. 4-d). The percentage of models' global sparsity increases the participation rate in parallel and both S-FL and subMFL have similar results. However, newly attended devices start to train a more accurate global model. For instance, the number of devices participating in the FL system increases by nearly 60 per cent for both S-FL and subMFL. However, due to transferred weights from $GM$, $SM$ shows good accuracy performance even before training. For the MNIST dataset, accuracy remains around 83 percent, until the $Threshold$ value reaches 0.8. For the FMNIST dataset, the accuracy maintains higher than 60 percent, until the $Threshold$ value reaches 0.7.

Comparing participation performance, assume that a middle-level sparse model is selected as a global model in S-FL. All devices will start to train a model with Threshold = 0.5 and random weights. Based on Fig. 4-b and Fig. 4-d, participation of devices will be under 45 per cent and higher computation capacity devices will train a model with higher learning constraints, in order to accommodate resource-constrained devices. On the other hand, subMFL provides each device to train and own optimal models by sharing sparse models in descending order. Those shared models are $SM$ generated by a pre-trained $GM$ which results in good accuracy, despite being compressed models. Fig. 4-b and Fig. 4-d show that subMFL increases participation up to 70 per cent.
\vspace{-0.6cm}
\subsection{Discussion} 
\vspace{-0.3cm}
Resource-constrained devices need smaller models to train and share their local models. In the case of selecting a small global model to share on the server side, due to the model's learning constraints, the model cannot generalise patterns in datasets. In the other case when a large global model is shared, resource-constrained devices are unable to train the model, due to computational capacity. This leads to bias in trained models and affects performance negatively. 

To provide dense DNNs to edge devices in heterogeneous environments, a flexible method should be utilised. State-of-the-art practice involves model pruning to compress these models for resource-limited devices. However, when the model pruning process is left to the device side, it results in extra energy consumption while they need to train their own local models. For this reason, it is necessary to generate methods to increase the heterogeneous device participation rate while pruning models on the server side. In this paper, we addressed this issue by serving compatible submodels to resource-constrained devices.

To sum up, only 10 per cent of data is utilised to train $GM$ due to 10 per cent of devices being capable to train it. However, results show that it is possible to generate compatible $SM$ via pruning $GM$ with different threshold values. Those $SM$ can be shared with resource-constrained devices as new global models to train. Thus, the number of participating devices increases, and since $SM$ owns tuned parameters from trained $GM$, those $SM$ start with good accuracy. 

The core idea of our work is to show that instead of selecting a random global model with a performance of around 10 per cent accuracy and distributing it to train (the S-FL approach), starting to train a dense global model and then pruning it to generate submodels is a useful approach, due to transferred pre-trained weights result in compressed submodels with a good accuracy performance. Even though new training rounds begin, those submodels can be served to resource-constrained devices that need to participate with a reasonable starting accuracy. Thus, at the end of the process, each device owns the optimal trainable model. 
\vspace{-0.5cm}
\section{Conclusion and Future Works}
\vspace{-0.25cm}
In this paper, we proposed subMFL which is a submodel generation technique using model pruning to increase the participation of heterogeneous devices for federated learning. This is done without a need for prior information on devices' hardware/computing capabilities or determining a specific pruning ratio. In our approach, a dense model is distributed to all devices in the system for training. Then, the trained model is pruned gradually in the server without the need for a data sample, so as to generate a set of submodels. Those submodels are shared with resource-constrained devices to train as compatible sparsed models to raise participation numbers. Also, since sparsed submodels hold tuned weights from the trained dense model, they have satisfactory accuracy even before training, despite being compressed. 

Future work could address more in-depth theoretical research regarding improving submodels used as global models that are trained with different device groups by combining their parameters to increase models' generality. Additionally, advanced compression methods can be used to reduce communication overhead in addition to tackling device heterogeneity. 
\vspace{-0.5cm}
\section*{Acknowledgment}
\vspace{-0.35cm}
This project has received funding from RE-ROUTE Project, the European Union’s Horizon Europe
research and innovation programme under the Marie Skłodowska-Curie grant
agreement No 101086343.
%
%
%
%

\end{document}